\documentclass[11pt]{article}

\usepackage[a4paper,margin=2.5cm]{geometry}
\usepackage{amsmath,amssymb,amsthm,bm}
\usepackage{booktabs,array}
\usepackage{graphicx}
\graphicspath{{figures/}}
\usepackage{algorithm}
\usepackage{algpseudocode}
\usepackage{float}
\usepackage[hidelinks]{hyperref}
\usepackage{url}
\hypersetup{
  unicode=true,
  pdftitle={Constructing Structured Decision Sources for Consensus-Based Pseudo-Label Learning},
  pdfauthor={Long Wang}
}

\newtheorem{definition}{Definition}

\title{Constructing Structured Decision Sources for Consensus-Based Pseudo-Label Learning}
\author{Long Wang\\
  {\small\href{mailto:alueas@126.com}{\texttt{alueas@126.com}}}\\
  {\small\href{https://orcid.org/0009-0006-8371-1562}{ORCID: 0009-0006-8371-1562}}}
\date{}

\begin{document}
\pagestyle{plain}
\maketitle

\begin{abstract}
Consensus can make pseudo-label learning more reliable, but only when its predictors contribute genuinely different evidence. Multiple models that repeat the same boundary provide additional votes without additional information. We address this problem by constructing decision sources through controlled changes to within-class structure. Starting from a shared graph representation, we vary center granularity and neighborhood mixing, reproduce each resulting source to test its stability, and select a complementary subset using node-pair co-assignment. Unanimous predictions from the selected sources are then ranked for student training. On the public fixed splits of Cora, CiteSeer, and PubMed, evaluated with five random seeds, the constructed sources improve fixed-budget training pseudo-label precision by 1.19--4.39 percentage points over three conventionally initialized GCN sources. Under matched structural filters, three-source consensus is more precise than each constituent source in all 45 dataset--slot--seed comparisons. The gains are strongest in pseudo-label quality: downstream accuracy remains competitive but does not lead on every dataset. These results identify source construction---rather than model count alone---as an important design problem for consensus-based pseudo-label learning.

\medskip
\noindent\textbf{Keywords:} graph semi-supervised learning; decision-source construction; multi-source consensus; near-center filtering; pseudo-label reliability
\end{abstract}

\section{Introduction}

Graph neural networks can exploit both node attributes and topology, yet their performance still depends on scarce labeled data. Pseudo-labeling extends supervision to unlabeled nodes, but an early error can be reinforced when the model later trains on its own prediction; message passing may then spread that error beyond the original node. Reliable pseudo-label admission is therefore a central problem in graph self-training \cite{kipf2017gcn}.

Most work controls pseudo-label noise after predictions have been produced, using confidence schedules, training dynamics, prototypes, or agreement among models \cite{cascante2021curriculum,zhang2021flexmatch,pei2024memory,nassar2023protocon,odonnat2024leveraging,to2025diverse}. Agreement is useful only if the participating predictors fail in different ways. Random initializations may instead recover nearly identical boundaries or introduce unstable variation. The unresolved question is therefore upstream of voting: how can a single graph yield decision sources that are reproducible enough to trust and different enough to cross-check one another?

DiPat demonstrates that one graph can support multiple decision patterns and that their agreement can guide pseudo-label selection \cite{liu2026dipat}. We study the preceding design problem: constructing, verifying, and selecting the sources that enter consensus. This distinction matters because apparent diversity can arise from transient optimization noise rather than a stable change in structural criterion.

Our framework varies the number of within-class centers $k$ and the mixing coefficient $\beta$ between node and neighborhood representations. Each intervention produces a candidate source from a shared warm start. Reproduction training tests whether its induced partition persists, and node-pair co-assignment measures whether two sources impose meaningfully different partitions without relying on permutable center identifiers. We select three stable, complementary sources from four candidates and rank their unanimous pseudo-labels by the weakest supporting confidence. A training-calibrated near-center gate provides a separate precision--coverage diagnostic; student training uses the ranked consensus labels directly.

Experiments on Cora, CiteSeer, and PubMed compare the constructed sources with a supervised GCN, a graph-only DiPat adaptation, three conventionally initialized GCN sources, and matched single-source controls. The constructed sources improve fixed-budget pseudo-label precision on all three datasets, and consensus improves precision over every matched constituent source. Downstream Accuracy and Macro-F1 remain competitive but are not uniformly superior, separating the quality of the admitted evidence from the final task outcome.

Our contributions are as follows:
\begin{itemize}
  \item We formulate source construction as part of the consensus-learning problem and instantiate it through controlled within-class granularity and neighborhood mixing.
  \item We connect source reproducibility and complementarity to consensus utility through co-assignment probes, matched single-source controls, and random-source replacement.
  \item Across three citation graphs, the resulting sources consistently improve pseudo-label evidence while revealing the accompanying coverage and downstream-performance tradeoffs.
\end{itemize}

\section{Related Work}

\subsection{Pseudo-label admission, correction, and reliability}

Pseudo-label admission methods determine when and how strongly model predictions should enter training. Fixed confidence rules \cite{sohn2020fixmatch}, curricula and dynamic thresholds \cite{cascante2021curriculum,xu2021dash}, class-adaptive thresholds \cite{zhang2021flexmatch,wang2023freematch}, and soft weighting \cite{chen2023softmatch} all manage the quantity--quality tradeoff after a predictor has produced its scores. Our question precedes admission: which predictors should supply those scores to a consensus rule?

Pseudo-labels can also be improved through teacher feedback, complementary classifiers, prototype consistency, or temporal disagreement \cite{pham2021meta,nassar2021semco,nassar2023protocon,pei2024memory}. These approaches refine supervision using information already present in the training process. We instead treat the generation and validation of differentiated decision signals as an explicit design problem.

\subsection{Structure and noise control in graph semi-supervised learning}

Graph semi-supervised methods exploit structure through clustering and self-training \cite{sun2020m3s}, auxiliary manifold mixing \cite{verma2021graphmix}, stochastic propagation and consistency \cite{feng2020grand,wang2020gsnn}, or explicit correction and smoothing \cite{huang2021correct}. They show that graph structure can regularize predictions, but generally use structural variation to strengthen one learner rather than to construct a set of comparable sources.

Robust graph learning further addresses noisy labels and topology through graph reconstruction, correction, and topology-aware sample selection \cite{dai2021nrgnn,dai2022rsgnn,li2023graphcleaner,wu2024tss,alchihabi2023robust}. Our intervention has a different role: center granularity and neighborhood mixing deliberately create candidate sources, and a shared co-assignment probe distinguishes reproducible structural variation from incidental optimization noise.

\subsection{Multi-source consensus, diversity, and structured decision sources}

Agreement models and diverse ensembles use cross-model consistency to control unlabeled supervision \cite{stretcu2019gam,to2025diverse,odonnat2024leveraging,zhou2018diverse}. Their gains depend on exploitable complementarity, not the number of members alone. This observation motivates treating diversity as something to construct and verify before voting.

DiPat is the closest comparison because it derives complementary decision patterns from one graph \cite{liu2026dipat}. It begins with predefined patterns and studies their use in consensus; we study how candidate sources can be generated, distinguished, and accepted before fusion. Structural interventions create the candidates, co-assignment makes their partitions comparable, and reproduction tests whether those partitions persist.

\section{Method}

\begin{figure}[H]
\centering
\includegraphics[width=\linewidth]{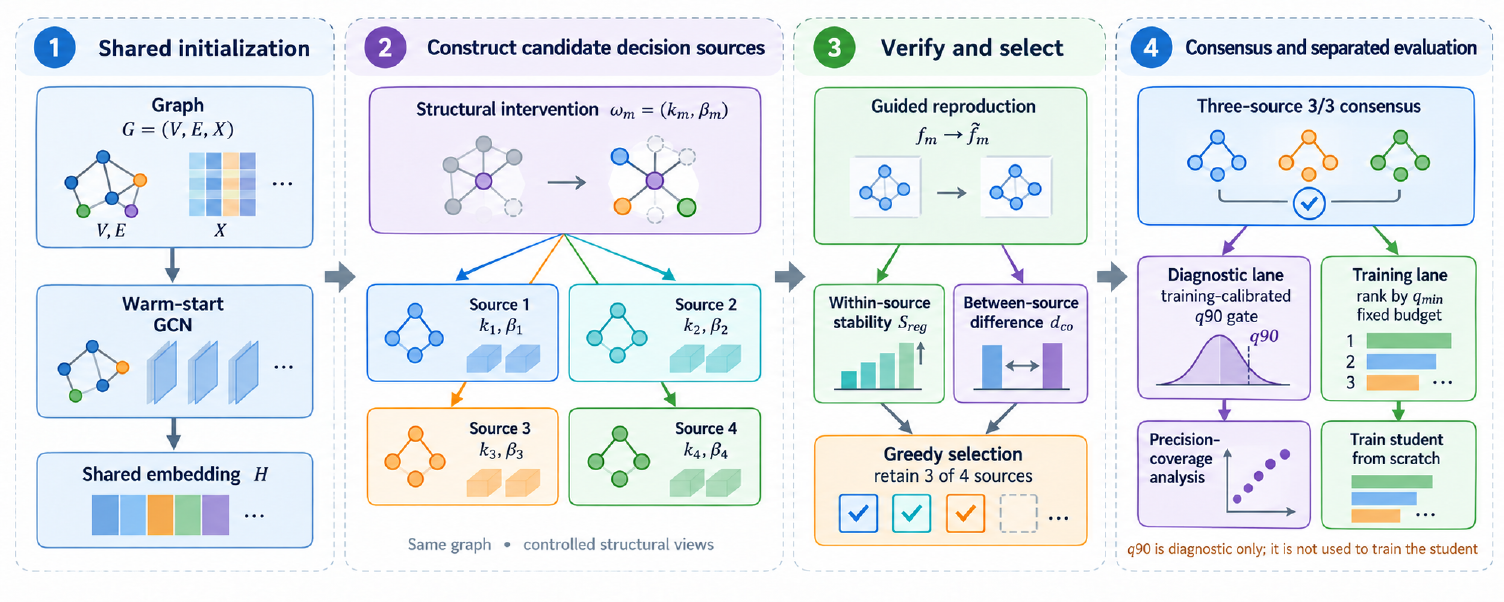}
\caption{Overview of structured decision-source construction. Structural interventions $(k,\beta)$ generate four candidates from a shared warm-start representation. Reproduction stability and between-source co-assignment differences select three sources. Their consensus candidates enter either the q90 quality diagnostic or minimum-confidence ranking for student training; the q90 gate is not used in student training.}
\label{fig:overall}
\end{figure}

Figure~\ref{fig:overall} summarizes the pipeline from a shared warm start through source construction and reproduction to consensus use. Controlled structural intervention precedes consensus learning, and the q90 quality diagnostic is kept separate from the student-training branch.

\subsection{Problem definition and notation}

Given a graph $G=(V,E,X)$, a public training-node set $T\subset V$, and class set $\mathcal Y$, the candidate domain is $U=V\setminus T$. This is a transductive setting: $U$ includes validation, test, and other unlabeled nodes, but their ground-truth labels are not used in source construction or filtering. Let $\Omega=\{\omega_1,\ldots,\omega_M\}$ be the intervention set, where $\omega_m=(k_m,\beta_m)$ specifies the number of within-class centers and the neighborhood mixing coefficient. Training on the same data under different interventions yields
\begin{equation}
 f_m(G)=\{p_m(y\mid i;\omega_m):i\in V\},\qquad m=1,\ldots,M.
\end{equation}

\begin{definition}[Sample co-assignment probe]
Group nodes in $U$ by the class predicted by the shared warm-start model, and sample a set of within-group node pairs $\mathcal P$ under a fixed random seed. Let $a_m(i)$ denote the within-class center assigned to node $i$ by source $m$. For $p=(i,j)\in\mathcal P$, define
\begin{equation}
 C_m(p)=\mathbb I[a_m(i)=a_m(j)].
\end{equation}
At most $5\times10^4$ pairs are sampled per seed, with quotas allocated according to the number of available pairs in each predicted class. This Boolean representation records only whether two nodes are co-assigned and is invariant to center-label permutations.
\end{definition}

The structural difference between sources $m$ and $m'$ is
\begin{equation}
 d_{\mathrm{co}}(m,m')=
 \frac{1}{|\mathcal P|}\sum_{p\in\mathcal P}|C_m(p)-C_{m'}(p)|.
\end{equation}
Structure-guided curriculum reproduction of source $m$ produces $\widetilde f_m$ and $\widetilde C_m$. Its reproduction stability is
\begin{equation}
 s_{\mathrm{reg}}(m)=1-
 \frac{1}{|\mathcal P|}\sum_{p\in\mathcal P}
 |C_m(p)-\widetilde C_m(p)|.
\end{equation}
The two quantities describe between-source differences and within-source reproducibility on the same fixed node pairs; they are not direct semantic measurements of internal network representations.

\subsection{Why complementary support can help}

Let $H$ be the event that a candidate pseudo-label is correct, with prior $P(H)=\pi$. Let $E_m$ be the event that source $m$ supports that label.

For intuition, suppose that $E_1,\ldots,E_K$ are conditionally independent given $H$ or $\neg H$, and that every retained source provides informative support:
\begin{equation}
 \frac{P(E_m\mid H)}{P(E_m\mid\neg H)}\geq \lambda>1.
\end{equation}
The posterior odds after unanimous support then satisfy
\begin{align}
 \frac{P(H\mid E_{1:K})}{P(\neg H\mid E_{1:K})}
 &=\frac{\pi}{1-\pi}
   \prod_{m=1}^{K}\frac{P(E_m\mid H)}{P(E_m\mid\neg H)}\\
 &\geq \frac{\pi}{1-\pi}\lambda^K.
\end{align}
Equivalently,
\begin{equation}
 P(H\mid E_1,\ldots,E_K)\geq
 \frac{\pi\lambda^K}{1-\pi+\pi\lambda^K}.
\end{equation}
This calculation supplies intuition, not a model of our trained sources: they share a graph and training process, and the experiments do not establish conditional independence or a common likelihood ratio. Our empirical criterion is consequently functional. A construction is useful when, under matched voting and filtering, its sources improve the precision--coverage tradeoff and the quality of the labels used for training.

\subsection{Structured construction, reproduction, and consensus}

\begin{figure}[H]
\centering
\includegraphics[width=\linewidth]{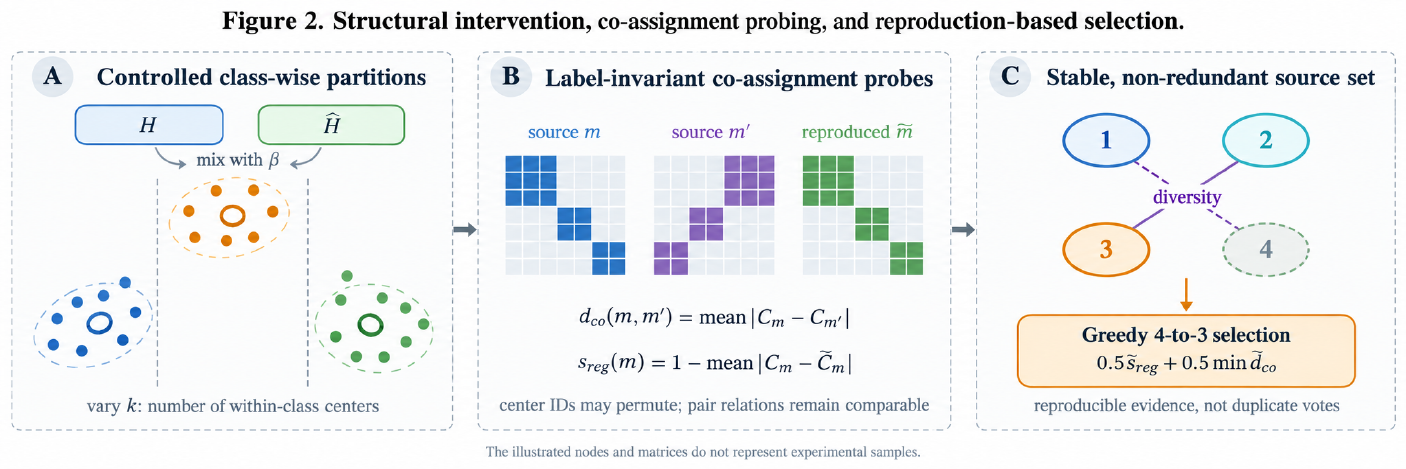}
\caption{Structural intervention, co-assignment probing, and reproduction-based selection. Varying $k$ and $\beta$ creates different within-class partitions. Co-assignment matrices remove center-label permutations, and within-source stability $s_{\mathrm{reg}}$ and between-source difference $d_{\mathrm{co}}$ select three stable, nonduplicate sources. The illustrated nodes and matrices do not represent experimental samples.}
\label{fig:mechanism}
\end{figure}

We first warm-start a GCN on labeled training nodes to obtain hidden representations $H$. For each $\omega=(k,\beta)$, we mix $H$ with symmetrically normalized neighborhood aggregation $\widehat A H$. Within each class, $k$ centers are fit using labeled nodes of that class together with non-training nodes predicted as that class with confidence at least 0.8. Non-training nodes use only predicted classes during center fitting and assignment. For every within-class center, contributing non-training nodes are ordered by distance to the center. Prefixes containing 50\%, 75\%, and 100\% of those nodes are successively combined with all labeled training nodes to train the source. The structure is then refit, and the same curriculum trains a reproduction model; differences between the initial and reproduced co-assignment vectors determine $s_{\mathrm{reg}}$. Figure~\ref{fig:mechanism} separates structural partitioning, co-assignment probing, and stable nonredundant selection. The default method imposes no hard minimum occupancy; Section~4.5 reports the corresponding ablation.

Co-assignment relations reveal redundancy among candidates. Rather than tune a post hoc hard threshold, we retain a fixed $K=3$ sources. Stability is min--max normalized over the four candidates; pairwise distances are min--max normalized over the six off-diagonal candidate pairs. A zero-range quantity is mapped to zero, and ties follow lexicographic order of $(k,\beta)$. The most reproducible source is selected first; further sources are greedily added using normalized stability and their minimum co-assignment distance from the selected set:
\begin{equation}
 g(m\mid\mathcal M^\star)=
 \frac{1}{2}\widetilde s_{\mathrm{reg}}(m)+
 \frac{1}{2}\min_{j\in\mathcal M^\star}\widetilde d_{\mathrm{co}}(m,j).
\end{equation}

For $i\in U$, each retained source predicts $\hat y_{m,i}$ with confidence $q_{m,i}=\max_y p_m(y\mid i)$. The unanimous three-source candidate set is
\begin{equation}
 \mathcal V_{3/3}=\{i\in U:\exists c\in\mathcal Y,
 \ \hat y_{m,i}=c\ \forall m\in\mathcal M^\star\}.
\end{equation}
Let $\hat y_i$ be the common label. In the \emph{pseudo-label quality} experiment, all nodes in $\mathcal V_{3/3}$ are evaluated before confidence ranking. For retained source $m$ and class $c$, let $d_{m,c}(i)$ be the distance from a node to its nearest center of class $c$. Using labeled training nodes only, define
\begin{equation}
 t_{m,c}=Q_{0.90}\bigl(\{d_{m,c}(i):i\in T,\ y_i=c\}\bigr),
 \qquad
 \mathcal S_{q90}=\{i\in\mathcal V_{3/3}:
 d_{m,\hat y_i}(i)\leq t_{m,\hat y_i},\ \forall m\in\mathcal M^\star\}.
\end{equation}
Quality predictions come from source models after six ranking rounds, whereas centers and their training-node q90 thresholds are fixed at the pre-ranking reproduction stage. For DiPat and the joint six-source setting, their voting sets replace $\mathcal V_{3/3}$ while the same three selected structural-source centers are used as an additional distance filter evaluated at the candidate's voted class. This gate supplies distance-filtering information, not another class vote, and is not a native DiPat score. A maximum normalized distance can be used offline to compute error-detection AUC, but it does not define node-level admission.

In the separate \emph{downstream task} experiment, the frozen source configuration undergoes six rounds of dynamic ranking. Consensus candidates are finally ordered by conservative confidence $q_{\min}(i)=\min_{m\in\mathcal M^\star}q_{m,i}$, and at most $\lceil0.675|U|\rceil$ nodes are used to train a student from scratch. The q90 gate is not used in this branch. The supervised and pseudo-label losses are
\begin{equation}
 \mathcal L=\mathcal L_{\mathrm{sup}}+\mathcal L_{\mathrm{pl}},\quad
 \mathcal L_{\mathrm{pl}}=-\frac{1}{|\mathcal S_u|}
 \sum_{(x,\hat y)\in\mathcal S_u}\log p(\hat y\mid x).
\end{equation}

\begin{algorithm}[htbp]
\caption{Structured source construction and separated quality and task evaluation}
\begin{algorithmic}[1]
\Require Graph, public training mask, intervention set $\Omega$, retained-source count $K=3$
\Ensure Voting and q90 quality statistics; independent student evaluation
\For{$\omega_m\in\Omega$}
  \State Train structured source $f_m$ and compute node-pair co-assignment vector $C_m$
  \State Build a curriculum from within-class center assignments and distances; retrain $\widetilde f_m$
  \State Compute reproduction stability $s_{\mathrm{reg}}(m)$
\EndFor
\State Select the source with highest $s_{\mathrm{reg}}$, then greedily add sources by $g(m\mid\mathcal M^\star)$ until $K$ are retained
\State Evaluate all non-training nodes admitted by voting and the subset retained by the training-calibrated q90 gate
\State Independently run source ranking, fix the final budget, and train a student from scratch
\State \Return Separated quality and task metrics
\end{algorithmic}
\end{algorithm}

If training one decision source costs $T_f$, explicitly constructing co-assignment matrices over $n$ nodes costs $O(Mn^2)$. With a fixed sample of $P$ pairs, relation probing costs $O(MP)$. We use four candidates and retain three throughout; increasing the number of sources is not treated as a cost-free performance guarantee.

\section{Experimental Design}

\subsection{Data, splits, and frozen configuration}

We use the Planetoid \texttt{public} fixed splits from PyTorch Geometric \cite{yang2016planetoid,fey2019pyg} and run seeds 0--4 on Cora, CiteSeer, and PubMed. Table~\ref{tab:data} reports dataset sizes verified against the data manifest. Each public split contains 20 training nodes per class. Validation and test nodes remain in the non-training candidate domain $U$, whose sizes are 2,568, 3,207, and 19,657. All datasets use only row-normalized original node features $X$; neither \texttt{cora\_2025.pt} nor external text or multimodal features is used. Seeds alter model initialization and training randomness but not the public masks.

\begin{table}[htbp]
\centering
\caption{Statistics and fixed masks for the three Planetoid public datasets. Edges are directed PyG edge entries.}
\label{tab:data}
\small
\begin{tabular}{lrrrrrr}
\toprule
Dataset & Nodes & Edge entries & Features & Classes & Train/val/test & $|U|$\\
\midrule
Cora & 2708 & 10556 & 1433 & 7 & 140/500/1000 & 2568\\
CiteSeer & 3327 & 9104 & 3703 & 6 & 120/500/1000 & 3207\\
PubMed & 19717 & 88648 & 500 & 3 & 60/500/1000 & 19657\\
\bottomrule
\end{tabular}
\end{table}

The source configuration was selected during Cora development and frozen for all three datasets without tuning on CiteSeer or PubMed test results. Four structural candidates are constructed by
\begin{equation}
 H_\beta=(1-\beta)H+\beta\widehat A H,\qquad
 (k,\beta)\in\{2,3\}\times\{0,0.5\}.
\end{equation}
Three are selected independently for each dataset and seed using reproduction stability and co-assignment distance, without labels beyond the public training set. Source curricula use 50\%, 75\%, and 100\% prefixes. Retained sources then undergo six ranking iterations of 10 epochs each; the admission ratio starts at 47.5\% and rises by four percentage points per round to 67.5\%. Source networks use hidden dimension 86 and dropout 0.39; students use hidden dimension 32 and dropout 0.5. The learning rate is 0.01 and weight decay is $5\times10^{-4}$. Final task budgets $\lceil0.675|U|\rceil$ are 1,734, 2,165, and 13,269 nodes. The 0.8 prediction-confidence condition used during source construction is distinct from both the q90 quality threshold and the student's ranking cutoff.

\subsection{Comparisons, metrics, and use of ground truth}

The main downstream controls are a supervised two-layer GCN with hidden dimension 64 and dropout 0.5, and a graph-only DiPat adaptation. The latter uses one shared marker input and a network with $3C$ outputs; it is not three independent models and does not include the original method's auxiliary \texttt{cat\_x} input \cite{liu2026dipat}. The source-replacement ablation uses three ordinary supervised GCNs with hidden dimension 86, dropout 0.39, and different initialization seeds. Voting, final ranking budget, and student initialization are aligned with our method, although total source-training computation is not matched to our guided, reproduction, and ranking stages. The joint six-source experiment performs only late voting after our sources and DiPat have been trained separately.

Pseudo-label precision $\mathrm{Prec}=|\{i\in S:\hat y_i=y_i\}|/|S|$ and coverage $\mathrm{Cov}=|S|/|U|$ are computed per seed and averaged over five seeds. We record counts of correct and incorrect nodes before and after q90 filtering, including the correct and incorrect pseudo-labels removed by the gate. This quality branch performs no confidence ranking and does not truncate to the task-training budget. The downstream branch reports only Test Accuracy and Macro-F1 of students trained from scratch. Training labels supervise models and calibrate q90; validation labels select checkpoints for both source models and students, even though validation nodes may also enter training with pseudo-labels; and test labels are used only for offline evaluation after filtering and training. This overlap in validation use is a limitation of the protocol. Raw files, code versions, node-level masks, and probabilities are retained in the experiment artifacts. All three datasets use the same implementation under Python 3.11.6, PyTorch 2.12.0+cu130, PyG 2.8.0.post1, and CUDA.

\section{Results}

\subsection{All voting candidates and the q90 structural gate}

Table~\ref{tab:quality} shows the central precision--coverage tradeoff. The q90 gate removes many candidates but raises conditional precision on every dataset, and stricter joint voting raises it further. Counts, precision, and coverage are arithmetic means of per-seed metrics; all filtered sets exclude labeled training nodes. For the DiPat and joint rows, q90 contributes structural information from our selected sources, so these rows evaluate external candidates under the same gate rather than native DiPat alone. In a DiPat view, an argmax outside that view's $C$-class output block abstains, and a joint label must receive the stated 4/6, 5/6, or 6/6 support.

\begin{table}[htbp]
\centering
\caption{Pseudo-label quality before and after q90 filtering (five-seed means). Precision and coverage are percentages. Removed correct/incorrect gives the numbers removed by q90. Within each dataset, bold marks the largest post-gate precision and the largest post-gate coverage.}
\label{tab:quality}
\scriptsize
\setlength{\tabcolsep}{2.5pt}
\resizebox{\linewidth}{!}{%
\begin{tabular}{llrrrrrr}
\toprule
Dataset & Voting rule & Before $N$ & Before prec. & After $N$ & After prec. & After cov. & Removed correct/incorrect\\
\midrule
Cora & \textbf{Structured sources 3/3} & 2473.6 & 84.27 & 1452.4 & 92.53 & \textbf{56.56} & 741.0/280.2\\
 & DiPat 3/3 + structural gate & 2437.4 & 83.23 & 1416.4 & 93.47 & 55.16 & 705.0/316.0\\
 & Joint 4/6 & 2388.4 & 85.36 & 1431.8 & 93.38 & 55.76 & 702.2/254.4\\
 & Joint 5/6 & 2298.8 & 86.87 & 1415.4 & 93.86 & 55.12 & 668.8/214.6\\
 & Joint 6/6 & 2208.8 & 88.21 & 1396.8 & \textbf{94.19} & 54.39 & 632.8/179.2\\
\midrule
CiteSeer & \textbf{Structured sources 3/3} & 3081.2 & 73.40 & 1322.0 & 85.86 & \textbf{41.22} & 1126.8/632.4\\
 & DiPat 3/3 + structural gate & 2981.0 & 73.16 & 1232.8 & 88.87 & 38.44 & 1086.0/662.2\\
 & Joint 4/6 & 2855.0 & 75.99 & 1247.6 & 88.71 & 38.90 & 1065.0/542.4\\
 & Joint 5/6 & 2738.8 & 77.34 & 1236.2 & 89.02 & 38.55 & 1019.2/483.4\\
 & Joint 6/6 & 2594.8 & 78.82 & 1222.8 & \textbf{89.34} & 38.13 & 952.4/419.6\\
\midrule
PubMed & \textbf{Structured sources 3/3} & 19283.2 & 79.36 & 14103.6 & 84.35 & \textbf{71.75} & 3410.8/1768.8\\
 & DiPat 3/3 + structural gate & 18746.8 & 79.68 & 13683.2 & 84.85 & 69.61 & 3329.0/1734.6\\
 & Joint 4/6 & 18535.0 & 80.61 & 13842.6 & 84.88 & 70.42 & 3194.6/1497.8\\
 & Joint 5/6 & 17972.2 & 81.42 & 13637.8 & 85.22 & 69.38 & 3013.8/1320.6\\
 & Joint 6/6 & 17378.0 & 82.25 & 13390.4 & \textbf{85.62} & 68.12 & 2830.2/1157.4\\
\bottomrule
\end{tabular}}
\end{table}

\begin{figure}[tbp]
\centering
\includegraphics[width=0.98\linewidth]{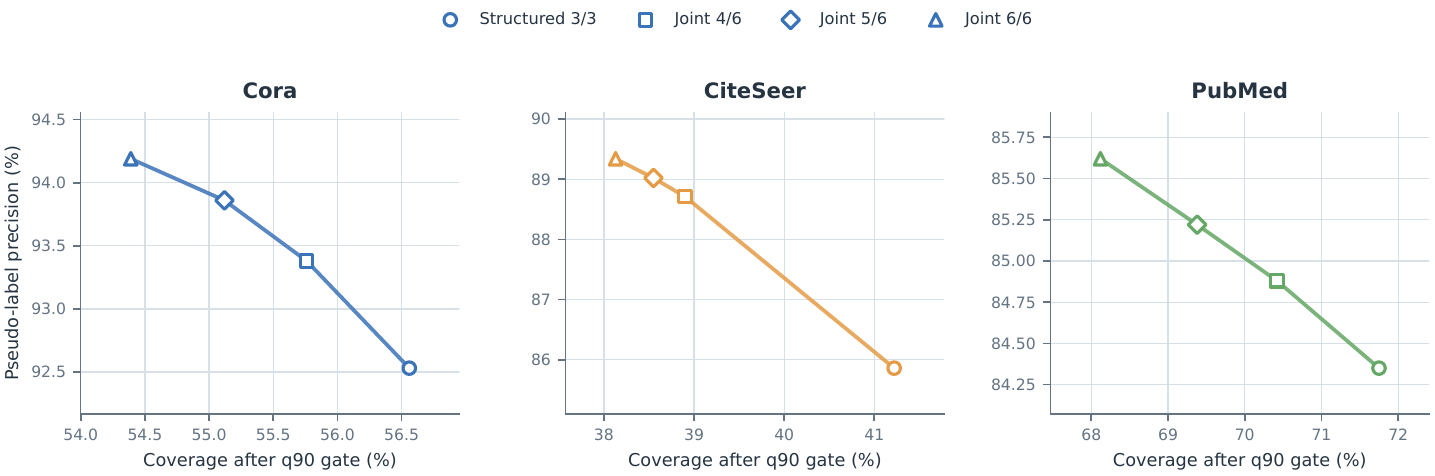}
\caption{Precision--coverage tradeoff after the common q90 structural gate (five-seed means). Lines connect adjacent voting strictness levels only to show direction and are not continuous threshold curves.}
\label{fig:precision-coverage}
\end{figure}

For structured 3/3 sources, q90 raises conditional precision by 8.26, 12.46, and 4.99 percentage points on Cora, CiteSeer, and PubMed, while coverage falls from 96.32\%, 96.08\%, and 98.10\% to 56.56\%, 41.22\%, and 71.75\%. Across seeds, post-gate precision has standard deviations of 0.74, 1.35, and 0.90 percentage points. Strict 6/6 consensus followed by the same gate reaches 94.19\%, 89.34\%, and 85.62\%, the highest precision in each dataset, with lower coverage of 54.39\%, 38.13\%, and 68.12\%. Figure~\ref{fig:precision-coverage} displays the common trend: stricter voting increases precision while reducing coverage. After the structural gate, DiPat candidates have higher precision than structured 3/3 candidates on all three datasets, showing that the gate can filter other candidate sources; conversely, the joint 6/6 result cannot be attributed to our three structural sources alone.

These results indicate that the training-calibrated near-center criterion reduces the risk of accepting incorrect labels. It should not be interpreted as evidence that output probabilities are calibrated. Error-detection AUC before gating is an auxiliary ranking diagnostic over continuous distance risk; node admission uses only the fixed q90 thresholds. Group counts and node-level votes, distances, thresholds, and pre-/post-gate masks are retained with the results for verification.

\subsection{Students trained from scratch}

The structured-source and graph-only DiPat methods in Table~\ref{tab:task} each select at most $\lceil0.675|U|\rceil$ pseudo-labels under the task protocol; the supervised GCN uses none. No method uses the q90 gate from Table~\ref{tab:quality} to train the student. Test Accuracy and Macro-F1 are five-seed means and should not be explained directly from post-gate precision.

\begin{table}[htbp]
\centering
\caption{Downstream results under the shared local protocol (Accuracy/F1 in percent; five-seed mean $\pm$ sample standard deviation). Students do not use the q90 gate. Within each dataset and metric, bold marks the largest mean; ties are included.}
\label{tab:task}
\small
\begin{tabular}{llrrr}
\toprule
Dataset & Method & Training pseudo-labels & Test Accuracy & Test Macro-F1\\
\midrule
Cora & Supervised GCN & 0 & $81.66\pm0.82$ & $80.51\pm0.82$\\
 & Graph-only DiPat adaptation & 1734 & $82.48\pm0.33$ & $81.31\pm0.47$\\
 & \textbf{Structured sources} & 1734 & $\bm{84.04\pm1.12}$ & $\bm{83.03\pm0.89}$\\
\midrule
CiteSeer & Supervised GCN & 0 & $71.46\pm0.66$ & $\bm{68.29\pm0.62}$\\
 & Graph-only DiPat adaptation & 2165 & $73.54\pm0.66$ & $67.26\pm0.61$\\
 & \textbf{Structured sources} & 2165 & $\bm{73.74\pm0.63}$ & $\bm{68.29\pm1.16}$\\
\midrule
PubMed & Supervised GCN & 0 & $79.26\pm0.32$ & $78.90\pm0.26$\\
 & Graph-only DiPat adaptation & 13269 & $\bm{80.26\pm0.67}$ & $\bm{79.73\pm0.60}$\\
 & \textbf{Structured sources} & 13269 & $79.94\pm0.54$ & $79.53\pm0.53$\\
\bottomrule
\end{tabular}
\end{table}

Structured sources achieve the highest mean Test Accuracy and Macro-F1 on Cora. On CiteSeer, Accuracy is 0.20 points above graph-only DiPat, while supervised GCN and structured sources have the same rounded F1. On PubMed, graph-only DiPat is slightly higher on both metrics. The experiment therefore does not support universal downstream superiority across graph datasets. The graph-only DiPat implementation is an adaptation under shared inputs rather than a full reproduction of the original method, which also uses auxiliary descriptions and its own feature and training protocol \cite{liu2026dipat}.

\subsection{Source-construction ablations and single-source controls}

Table~\ref{tab:ablation} replaces the complete construction pipeline with three ordinary supervised GCN sources of the same architecture and different initializations. The groups use the same 3/3 rule, q90 centers from the selected structural sources, final ranking quota, and student initialization. Ordinary GCN sources do not have within-class structural centers, so their post-gate values represent ordinary-source candidates plus our structural gate. The first two quality columns evaluate all voting candidates, whereas training pseudo-label precision is computed on the separately ranked, fixed task budget.

\begin{table}[htbp]
\centering
\caption{Replacement ablation against three conventionally initialized GCN sources (five-seed means, percent). Both post-gate rows use the same additional q90 structural gate; task training does not use the gate. Within each dataset and metric, bold marks the larger value; q90 coverage is not ranked because it trades off against precision.}
\label{tab:ablation}
\scriptsize
\setlength{\tabcolsep}{3pt}
\resizebox{\linewidth}{!}{%
\begin{tabular}{llrrrrr}
\toprule
Dataset & Decision sources & All-vote prec. & q90 prec./cov. & Training PL prec. & Test Acc. & Test Macro-F1\\
\midrule
Cora & \textbf{Structured sources} & 84.27 & \textbf{92.53}/56.56 & \textbf{93.04} & \textbf{84.04} & \textbf{83.03}\\
 & Three random-seed GCNs & \textbf{84.80} & 91.94/57.06 & 91.30 & 83.30 & 81.95\\
\midrule
CiteSeer & \textbf{Structured sources} & \textbf{73.40} & \textbf{85.86}/41.22 & \textbf{83.71} & \textbf{73.74} & 68.29\\
 & Three random-seed GCNs & 71.55 & 84.63/42.03 & 79.33 & 73.08 & \textbf{69.21}\\
\midrule
PubMed & \textbf{Structured sources} & 79.36 & \textbf{84.35}/71.75 & \textbf{88.18} & \textbf{79.94} & \textbf{79.53}\\
 & Three random-seed GCNs & \textbf{79.96} & 84.06/71.10 & 87.00 & 79.82 & 79.35\\
\bottomrule
\end{tabular}}
\end{table}

\begin{figure}[tbp]
\centering
\includegraphics[width=0.96\linewidth]{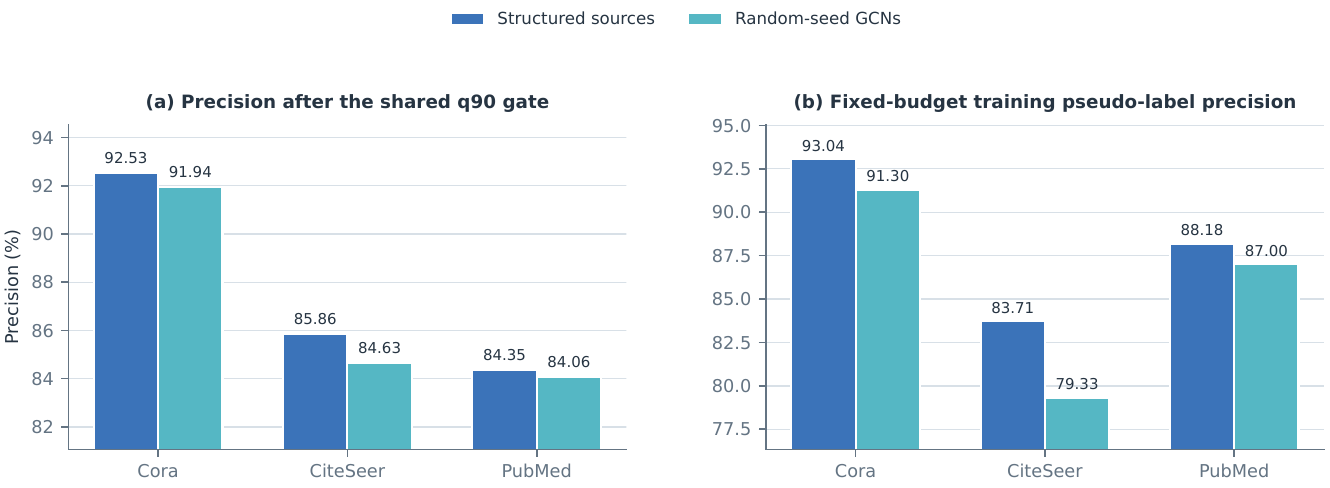}
\caption{Pseudo-label quality for structured sources and three random-seed GCNs (five-seed means). Left: conditional precision after the shared q90 structural gate. Right: fixed-budget training pseudo-label precision without q90 filtering. The panels evaluate different branches and should not be compared numerically across panels.}
\label{fig:source-bars}
\end{figure}

Under the same q90 gate, structured-source conditional precision exceeds the random-source control by 0.59, 1.23, and 0.29 percentage points on Cora, CiteSeer, and PubMed. Computed from unrounded per-seed aggregates, fixed-budget training pseudo-label precision improves by 1.74, 4.39, and 1.19 points; subtracting the displayed two-decimal means can differ by 0.01. As Figure~\ref{fig:source-bars} shows, this training-label improvement is positive for all five seeds on every dataset. Before gating, however, the random sources are slightly more precise on Cora and PubMed. Downstream gains are also mixed: the random sources have higher CiteSeer Macro-F1, and PubMed gains are only 0.12/0.18 points in Accuracy/F1. The comparison therefore supports a consistent pseudo-label-quality gain, while the unmatched multi-stage computation prevents attributing the full difference uniquely to $k$ or $\beta$.

A direct test of voting compares predictions from one retained source under its q90 gate with unanimous 3/3 labels under that same source's gate. Each cell in Table~\ref{tab:single} reports the latter minus the former in post-gate precision points. Source slots indicate selection order within a seed, not fixed $(k,\beta)$ settings. All five seeds in every slot and dataset yield positive gains of 0.25--0.80 points. Three-source voting removes some nodes relative to one source. In a separate comparison, moving from one matched gate to the full three-gate configuration raises conditional precision further at an additional coverage cost.

\begin{table}[htbp]
\centering
\caption{Post-gate precision gain of unanimous 3/3 voting over one source under the same single structural gate (percentage points; five-seed mean). Parentheses show seeds with positive gain.}
\label{tab:single}
\small
\begin{tabular}{lrrr}
\toprule
Dataset & Selected slot 0 & Selected slot 1 & Selected slot 2\\
\midrule
Cora & +0.61 (5/5) & +0.80 (5/5) & +0.42 (5/5)\\
CiteSeer & +0.66 (5/5) & +0.40 (5/5) & +0.38 (5/5)\\
PubMed & +0.34 (5/5) & +0.25 (5/5) & +0.32 (5/5)\\
\bottomrule
\end{tabular}
\end{table}

The matched-gate test uses predictions from post-ranking source models and centers from the pre-ranking reproduction stage, consistent with the historical q90 diagnostic in Table~\ref{tab:quality}. It is a selective filtering comparison rather than an equal-coverage threshold test.

\subsection{Structure as a difference probe}

Co-assignment is a statistical probe of whether selected sources impose different substructure criteria on the same node pairs. All datasets use the probe in Definition~1: for each seed, at most $5\times10^4$ shared pairs are sampled from non-training nodes within warm-start predicted classes, and Boolean center co-assignment vectors are compared at the reproduction stage. Table~\ref{tab:structure-probe} summarizes 15 reproduction-stability values and 15 pairwise co-assignment distances per dataset. CiteSeer and PubMed replay only the construction stage; selected source identifiers and node-level nearest-center distances match the original experiment cache, and ranking iterations and students are not rerun. Cora uses the existing records under the same definition.

\begin{table}[htbp]
\centering
\caption{Shared-pair structural probes for selected sources (five seeds; each statistic summarizes 15 values). Distances and stability lie in $[0,1]$.}
\label{tab:structure-probe}
\small
\begin{tabular}{lrrr}
\toprule
Dataset & Mean reproduction stability & Mean co-assignment distance & Distance range\\
\midrule
Cora & 0.833 & 0.222 & 0.140--0.303\\
CiteSeer & 0.814 & 0.263 & 0.128--0.412\\
PubMed & 0.918 & 0.234 & 0.088--0.329\\
\bottomrule
\end{tabular}
\end{table}

All three datasets show nonzero co-assignment differences, and selected source identifiers vary across seeds. A separate \emph{distance-risk} probe gives mean pairwise correlations of normalized near-center distance ratios of 0.799 and 0.811 for CiteSeer and PubMed, with q90-gate disagreement rates of 0.188 and 0.140 among structured 3/3 candidates. This is a different statistic from co-assignment distance. Because the sampled node-pair populations differ by dataset, distance magnitudes are not a direct ranking of source-construction quality; utility is assessed by the filtering and task controls in Tables~\ref{tab:quality}--\ref{tab:single}.

\subsection{Negative and boundary ablations}

Adding a minimum center-occupancy anti-collapse constraint to the frozen Cora configuration reduces five-seed Test Accuracy from 84.04\% to 83.44\% and Macro-F1 from 83.03\% to 82.61\%. This constraint changes the construction trajectory but does not improve mean task performance under the tested form and strength. Its pseudo-label metrics use the student's final ranking budget and are not mixed with the all-vote/q90 quantities in Table~\ref{tab:quality}.

A compound diagnostic shuffles center assignments and replaces curriculum distances with random values. Its student Accuracy is 83.38\%, 73.84\%, and 80.06\% on Cora, CiteSeer, and PubMed, with post-q90 precision of 92.46\%, 85.51\%, and 84.46\%. Because the diagnostic changes assignment and curriculum order together while retaining the original centers for gating, it does not isolate either component. The absence of a consistent downstream loss cautions against assigning a separate causal effect to every stage of the pipeline.

\section{Discussion and Limitations}

The most direct evidence for source construction comes from the matched functional controls. Consensus improves precision over every constituent source under the same gate, and the fixed-budget labels used for training are more precise than those from three random sources on all three datasets. Co-assignment distance explains how the selected sources differ, but the filtering comparisons establish whether that difference is useful. The benefit is concentrated in pseudo-label quality: coverage is lower, source-training computation is unmatched, and downstream results remain mixed.

The q90 diagnostic exposes the cost of selective reliability. It raises conditional precision but removes many correct labels, most sharply on CiteSeer, where coverage falls to 41.22\%. Its effect on DiPat and six-source candidates also shows that the learned geometry can filter labels proposed elsewhere. We therefore interpret q90 as a selective structural diagnostic, not as probability calibration or part of student training.

The theoretical bound assumes conditional independence and a likelihood-ratio lower bound, neither of which is verified for sources sharing a graph and training process. The configuration was developed on Cora, where test results were viewed during development. CiteSeer and PubMed therefore provide the more informative frozen-parameter checks, though they add only two citation graphs of the same type. Five seeds share the same public split and do not replace multi-split or out-of-domain validation. Validation nodes may be pseudo-labeled for training and are also used for checkpoint selection; test labels are used only for final accounting. Reported Accuracy and F1 values are descriptive means under the fixed protocol rather than significance claims.

\section{Conclusion}

We presented a pipeline that constructs consensus sources through controlled within-class interventions and selects them by reproducibility and complementarity. Across three citation graphs, the resulting sources provide more precise fixed-budget training labels than conventional random-seed sources, and their consensus is consistently more precise than matched constituent predictions. The downstream gains vary by dataset, so the evidence supports structured source construction as a way to improve pseudo-label admission---not as a universal route to the highest test accuracy.

\bibliographystyle{unsrt}
\bibliography{AB_source_learning_en}

@article{liu2026dipat,
  author  = {Kai Liu and Long Wang},
  title   = {Differentiated Information Mining: Semi-Supervised Graph Learning with Independent Patterns},
  journal = {Mathematics},
  volume  = {14},
  number  = {2},
  pages   = {279},
  year    = {2026},
  doi     = {10.3390/math14020279}
}

@inproceedings{kipf2017gcn,
  author    = {Thomas N. Kipf and Max Welling},
  title     = {Semi-Supervised Classification with Graph Convolutional Networks},
  booktitle = {International Conference on Learning Representations},
  year      = {2017},
  url       = {https://arxiv.org/abs/1609.02907}
}

@inproceedings{yang2016planetoid,
  author    = {Zhilin Yang and William W. Cohen and Ruslan Salakhutdinov},
  title     = {Revisiting Semi-Supervised Learning with Graph Embeddings},
  booktitle = {Proceedings of the 33rd International Conference on Machine Learning},
  year      = {2016},
  url       = {https://proceedings.mlr.press/v48/yanga16.html}
}

@article{fey2019pyg,
  author  = {Matthias Fey and Jan Eric Lenssen},
  title   = {Fast Graph Representation Learning with PyTorch Geometric},
  journal = {arXiv preprint arXiv:1903.02428},
  year    = {2019},
  url     = {https://arxiv.org/abs/1903.02428}
}

@inproceedings{to2025diverse,
  author    = {Nguyen Nhat Minh To and Paul F. R. Wilson and Viet Nguyen and Mohamed Harmanani and Michael Cooper and Fahimeh Fooladgar and Purang Abolmaesumi and Parvin Mousavi and Rahul Krishnan},
  title     = {Diverse Prototypical Ensembles Improve Robustness to Subpopulation Shift},
  booktitle = {Proceedings of the 42nd International Conference on Machine Learning},
  series    = {Proceedings of Machine Learning Research},
  volume    = {267},
  pages     = {59761--59783},
  year      = {2025},
  url       = {https://proceedings.mlr.press/v267/to25a.html}
}

@inproceedings{pei2024memory,
  author    = {Hongbin Pei and Yuheng Xiong and Pinghui Wang and Jing Tao and Jialun Liu and Huiqi Deng and Jie Ma and Xiaohong Guan},
  title     = {Memory Disagreement: A Pseudo-Labeling Measure from Training Dynamics for Semi-supervised Graph Learning},
  booktitle = {Proceedings of the ACM Web Conference 2024},
  pages     = {434--445},
  year      = {2024},
  doi       = {10.1145/3589334.3645398}
}

@inproceedings{nassar2023protocon,
  author    = {Islam Nassar and Munawar Hayat and Ehsan Abbasnejad and Hamid Rezatofighi and Gholamreza Haffari},
  title     = {ProtoCon: Pseudo-Label Refinement via Online Clustering and Prototypical Consistency for Efficient Semi-Supervised Learning},
  booktitle = {Proceedings of the IEEE/CVF Conference on Computer Vision and Pattern Recognition},
  pages     = {11641--11650},
  year      = {2023}
}

@inproceedings{odonnat2024leveraging,
  author    = {Ambroise Odonnat and Vasilii Feofanov and Ievgen Redko},
  title     = {Leveraging Ensemble Diversity for Robust Self-Training in the Presence of Sample Selection Bias},
  booktitle = {Proceedings of the 27th International Conference on Artificial Intelligence and Statistics},
  series    = {Proceedings of Machine Learning Research},
  volume    = {238},
  pages     = {595--603},
  year      = {2024},
  url       = {https://proceedings.mlr.press/v238/odonnat24a.html}
}

@inproceedings{zhou2018diverse,
  author    = {Tianyi Zhou and Shengjie Wang and Jeff A. Bilmes},
  title     = {Diverse Ensemble Evolution: Curriculum Data-Model Marriage},
  booktitle = {Advances in Neural Information Processing Systems},
  volume    = {31},
  year      = {2018}
}

@inproceedings{cascante2021curriculum,
  author    = {Paola Cascante-Bonilla and Fuwen Tan and Yanjun Qi and Vicente Ordonez},
  title     = {Curriculum Labeling: Revisiting Pseudo-Labeling for Semi-Supervised Learning},
  booktitle = {Proceedings of the AAAI Conference on Artificial Intelligence},
  volume    = {35},
  pages     = {6912--6920},
  year      = {2021},
  doi       = {10.1609/aaai.v35i8.16852}
}

@inproceedings{zhang2021flexmatch,
  author    = {Bowen Zhang and Yidong Wang and Wenxin Hou and Hao Wu and Jindong Wang and Manabu Okumura and Takahiro Shinozaki},
  title     = {FlexMatch: Boosting Semi-Supervised Learning with Curriculum Pseudo Labeling},
  booktitle = {Advances in Neural Information Processing Systems},
  volume    = {34},
  year      = {2021}
}

@inproceedings{sohn2020fixmatch,
  author    = {Kihyuk Sohn and David Berthelot and Nicholas Carlini and Zizhao Zhang and Han Zhang and Colin A. Raffel and Ekin Dogus Cubuk and Alexey Kurakin and Chun-Liang Li},
  title     = {FixMatch: Simplifying Semi-Supervised Learning with Consistency and Confidence},
  booktitle = {Advances in Neural Information Processing Systems},
  volume    = {33},
  year      = {2020},
  url       = {https://proceedings.neurips.cc/paper/2020/hash/06964dce9addb1c5cb5d6e3d9838f733-Abstract.html}
}

@inproceedings{xu2021dash,
  author    = {Yi Xu and Lei Shang and Jinxing Ye and Qi Qian and Yu-Feng Li and Baigui Sun and Hao Li and Rong Jin},
  title     = {Dash: Semi-Supervised Learning with Dynamic Thresholding},
  booktitle = {Proceedings of the 38th International Conference on Machine Learning},
  series    = {Proceedings of Machine Learning Research},
  volume    = {139},
  pages     = {11525--11536},
  year      = {2021},
  url       = {https://proceedings.mlr.press/v139/xu21e.html}
}

@inproceedings{wang2023freematch,
  author    = {Yidong Wang and Hao Chen and Qiang Heng and Wenxin Hou and Yue Fan and Zhen Wu and Jindong Wang and Marios Savvides and Takahiro Shinozaki and Bhiksha Raj and Bernt Schiele and Xing Xie},
  title     = {FreeMatch: Self-Adaptive Thresholding for Semi-Supervised Learning},
  booktitle = {International Conference on Learning Representations},
  year      = {2023},
  url       = {https://openreview.net/forum?id=PDrUPTXJI_A}
}

@inproceedings{chen2023softmatch,
  author    = {Hao Chen and Ran Tao and Yue Fan and Yidong Wang and Jindong Wang and Bernt Schiele and Xing Xie and Bhiksha Raj and Marios Savvides},
  title     = {SoftMatch: Addressing the Quantity-Quality Trade-Off in Semi-Supervised Learning},
  booktitle = {International Conference on Learning Representations},
  year      = {2023},
  url       = {https://openreview.net/forum?id=ymt1zQXBDiF}
}

@inproceedings{pham2021meta,
  author    = {Hieu Pham and Zihang Dai and Qizhe Xie and Quoc V. Le},
  title     = {Meta Pseudo Labels},
  booktitle = {Proceedings of the IEEE/CVF Conference on Computer Vision and Pattern Recognition},
  pages     = {11557--11568},
  year      = {2021},
  url       = {https://openaccess.thecvf.com/content/CVPR2021/html/Pham_Meta_Pseudo_Labels_CVPR_2021_paper.html}
}

@inproceedings{nassar2021semco,
  author    = {Islam Nassar and Samitha Herath and Ehsan Abbasnejad and Wray Buntine and Gholamreza Haffari},
  title     = {All Labels Are Not Created Equal: Enhancing Semi-Supervision via Label Grouping and Co-Training},
  booktitle = {Proceedings of the IEEE/CVF Conference on Computer Vision and Pattern Recognition},
  pages     = {7241--7250},
  year      = {2021},
  url       = {https://openaccess.thecvf.com/content/CVPR2021/html/Nassar_All_Labels_Are_Not_Created_Equal_Enhancing_Semi-Supervision_via_Label_CVPR_2021_paper.html}
}

@inproceedings{sun2020m3s,
  author    = {Ke Sun and Zhouchen Lin and Zhanxing Zhu},
  title     = {Multi-Stage Self-Supervised Learning for Graph Convolutional Networks on Graphs with Few Labeled Nodes},
  booktitle = {Proceedings of the AAAI Conference on Artificial Intelligence},
  volume    = {34},
  number    = {4},
  pages     = {5892--5899},
  year      = {2020},
  doi       = {10.1609/aaai.v34i04.6048}
}

@inproceedings{verma2021graphmix,
  author    = {Vikas Verma and Meng Qu and Kenji Kawaguchi and Alex Lamb and Yoshua Bengio and Juho Kannala and Jian Tang},
  title     = {GraphMix: Improved Training of {GNN}s for Semi-Supervised Learning},
  booktitle = {Proceedings of the AAAI Conference on Artificial Intelligence},
  volume    = {35},
  number    = {11},
  pages     = {10024--10032},
  year      = {2021},
  doi       = {10.1609/aaai.v35i11.17203}
}

@inproceedings{feng2020grand,
  author    = {Wenzheng Feng and Jie Zhang and Yuxiao Dong and Yu Han and Huanbo Luan and Qian Xu and Qiang Yang and Evgeny Kharlamov and Jie Tang},
  title     = {Graph Random Neural Networks for Semi-Supervised Learning on Graphs},
  booktitle = {Advances in Neural Information Processing Systems},
  volume    = {33},
  year      = {2020},
  url       = {https://proceedings.neurips.cc/paper/2020/hash/fb4c835feb0a65cc39739320d7a51c02-Abstract.html}
}

@inproceedings{wang2020gsnn,
  author    = {Haibo Wang and Chuan Zhou and Xin Chen and Jia Wu and Shirui Pan and Jilong Wang},
  title     = {Graph Stochastic Neural Networks for Semi-Supervised Learning},
  booktitle = {Advances in Neural Information Processing Systems},
  volume    = {33},
  year      = {2020},
  url       = {https://proceedings.neurips.cc/paper/2020/hash/e586a4f55fb43a540c2e9dab45e00f53-Abstract.html}
}

@inproceedings{huang2021correct,
  author    = {Qian Huang and Horace He and Abhay Singh and Ser-Nam Lim and Austin R. Benson},
  title     = {Combining Label Propagation and Simple Models Out-Performs Graph Neural Networks},
  booktitle = {International Conference on Learning Representations},
  year      = {2021},
  url       = {https://openreview.net/forum?id=8E1-f3VhX1o}
}

@inproceedings{dai2021nrgnn,
  author    = {Enyan Dai and Charu Aggarwal and Suhang Wang},
  title     = {{NRGNN}: Learning a Label Noise Resistant Graph Neural Network on Sparsely and Noisily Labeled Graphs},
  booktitle = {Proceedings of the 27th ACM SIGKDD Conference on Knowledge Discovery and Data Mining},
  pages     = {227--236},
  year      = {2021},
  doi       = {10.1145/3447548.3467364}
}

@inproceedings{dai2022rsgnn,
  author    = {Enyan Dai and Wei Jin and Hui Liu and Suhang Wang},
  title     = {Towards Robust Graph Neural Networks for Noisy Graphs with Sparse Labels},
  booktitle = {Proceedings of the Fifteenth ACM International Conference on Web Search and Data Mining},
  pages     = {181--191},
  year      = {2022},
  doi       = {10.1145/3488560.3498408}
}

@inproceedings{li2023graphcleaner,
  author    = {Yuwen Li and Miao Xiong and Bryan Hooi},
  title     = {GraphCleaner: Detecting Mislabelled Samples in Popular Graph Learning Benchmarks},
  booktitle = {Proceedings of the 40th International Conference on Machine Learning},
  series    = {Proceedings of Machine Learning Research},
  volume    = {202},
  pages     = {20195--20209},
  year      = {2023},
  url       = {https://proceedings.mlr.press/v202/li23ai.html}
}

@inproceedings{wu2024tss,
  author    = {Yuhao Wu and Jiangchao Yao and Xiaobo Xia and Jun Yu and Ruxin Wang and Bo Han and Tongliang Liu},
  title     = {Mitigating Label Noise on Graphs via Topological Sample Selection},
  booktitle = {Proceedings of the 41st International Conference on Machine Learning},
  series    = {Proceedings of Machine Learning Research},
  volume    = {235},
  pages     = {53944--53972},
  year      = {2024},
  url       = {https://proceedings.mlr.press/v235/wu24ae.html}
}

@inproceedings{alchihabi2023robust,
  author    = {Abdullah Alchihabi and Yuhong Guo},
  title     = {Learning Robust Graph Neural Networks with Limited Supervision},
  booktitle = {Proceedings of the 26th International Conference on Artificial Intelligence and Statistics},
  series    = {Proceedings of Machine Learning Research},
  volume    = {206},
  pages     = {8723--8733},
  year      = {2023},
  url       = {https://proceedings.mlr.press/v206/alchihabi23a.html}
}

@inproceedings{stretcu2019gam,
  author    = {Otilia Stretcu and Krishnamurthy Viswanathan and Dana Movshovitz-Attias and Emmanouil Antonios Platanios and Sujith Ravi and Andrew Tomkins},
  title     = {Graph Agreement Models for Semi-Supervised Learning},
  booktitle = {Advances in Neural Information Processing Systems},
  volume    = {32},
  year      = {2019},
  url       = {https://proceedings.neurips.cc/paper/2019/hash/4772c1b987f1f6d8c9d4ef0f3b764f7a-Abstract.html}
}
\end{document}